\documentclass[letterpaper, 10 pt, conference]{ieeeconf}  

\IEEEoverridecommandlockouts                              

\input{bib_short.def}

\usepackage{graphics} 
\usepackage{epsfig} 
\usepackage{mathptmx} 
\usepackage{times} 
\usepackage{amsmath} 
\usepackage{amssymb}  
\usepackage{booktabs}
\usepackage{multirow} 
\usepackage{makecell}
\usepackage{tikz}
\usetikzlibrary{arrows.meta, positioning,fit, calc,backgrounds}
\usepackage{url}
\usepackage[hidelinks]{hyperref}

\renewcommand{\thefootnote}{\fnsymbol{footnote}}

\usepackage{xcolor,colortbl}
\definecolor{Gray}{gray}{0.85}
\definecolor{LightCyan}{rgb}{0.88,1,1}

\newcolumntype{a}{>{\columncolor{Gray}}c}
\newcolumntype{b}{>{\columncolor{white}}c}

\newlength\barwidth  
\newlength\barheight 

\definecolor{colA}{RGB}{31,119,180}
\definecolor{colB}{RGB}{255,127,14}
\definecolor{colC}{RGB}{44,160,44}

\title{\LARGE \bf
Driver Behavior Estimation at Signalized Intersections Using a Physics-Constrained Decision-Conditioned Autoregressive Transformer
}

\author{Mohammad Khoshkdahan$^{1}$, Pavel Laskov$^{2}$, and Alexey Vinel$^{1,3}$%
\thanks{$^{1}$Mohammad Khoshkdahan is a Ph.D. student and Alexey Vinel is a Professor at Karlsruhe Institute of Technology (KIT), Germany.
{\tt\small \href{mailto:mohammad.khoshkdahan@kit.edu}{mohammad.khoshkdahan@kit.edu}}}%
\thanks{$^{2}$Pavel Laskov is a Professor at the University of Liechtenstein, Vaduz, Liechtenstein.}%
\thanks{$^{3}$Alexey Vinel is also a Professor at Halmstad University, Sweden.}%
}

\begin{document}

\maketitle
\thispagestyle{empty}
\pagestyle{empty}


\begin{abstract}
Red-light violations and harsh braking at signalized intersections are major contributors to traffic accidents. This paper analyzes and predicts human driver decision-making and longitudinal trajectory behavior during traffic light signal transitions. We collected a diverse real-world dataset comprising 449 approach runs under varying speed and distance conditions. Vehicle motion was recorded using RTK-corrected GNSS with centimeter-level accuracy, and driver heart rate and multi-level comfort ratings were monitored. Spatial and temporal calibration ensured precise alignment between vehicle state and signal timing. Statistical analysis identifies required deceleration as the dominant single predictor of the stop--go decision, and heteroscedastic Gaussian modeling of peak deceleration reveals five empirical comfort ranges derived from human stopping behavior. Based on this insight, we propose a two-stage modeling framework. Stage 1 predicts the binary maneuver decision, and Stage 2 generates the longitudinal acceleration trajectory using a decision-conditioned autoregressive Transformer with physics constraints, including target-state conditioning and jerk limits. The proposed architecture outperforms baseline methods and achieves 0.49~m/s$^2$ acceleration MAE and 0.62~m distance MAE. It also estimates the future stopping-comfort level of the human driver from a single yellow-onset snapshot. Qualitative results demonstrate realistic human-like braking behavior. The dataset and source code are publicly available.\footnote{\href{https://github.com/mohammadkhsh/driver-2s-transformer}{https://github.com/mohammadkhsh/driver-2s-transformer}}
\end{abstract}

\renewcommand{\thefootnote}{\arabic{footnote}}


\section{Introduction}
\label{sec:introduction}

Signalized intersections are critical points in urban traffic networks where timing and driver response directly influence safety. Inappropriate reactions during green-to-yellow transitions create two major risks. Abrupt braking may cause rear-end collisions, while late crossing during red can lead to severe intersection crashes. Drivers must therefore either decelerate smoothly and stop before the signal or proceed safely through the intersection under yellow. Because this decision involves uncertainty, perception, and vehicle dynamics under time pressure, accurate modeling requires joint understanding of both maneuver choice and trajectory execution. 

Predictive models that capture both the stop–go decision and the full braking trajectory are essential for reducing aggressive stopping and red-light violations. Such models can support adaptive signal timing or vehicle-to-infrastructure (V2I) systems that anticipate unsafe maneuvers \cite{xu2017v2i}. If the driver’s physiological response and perceived braking stress can be estimated from vehicle dynamics together with contextual factors such as demographics, signal timing, and environmental conditions, interventions can be timed more precisely. Speed adaptation can begin early and smoothly, potentially even while the signal is still green. This reduces the risk of late harsh braking that leads to violations and avoids premature deceleration that may cause rear-end conflicts. Accurate trajectory prediction therefore enables both safety and comfort to be improved in a coordinated manner.

\begin{figure}
\vspace{0.05cm}
    \centering
    \includegraphics[width=1\linewidth]{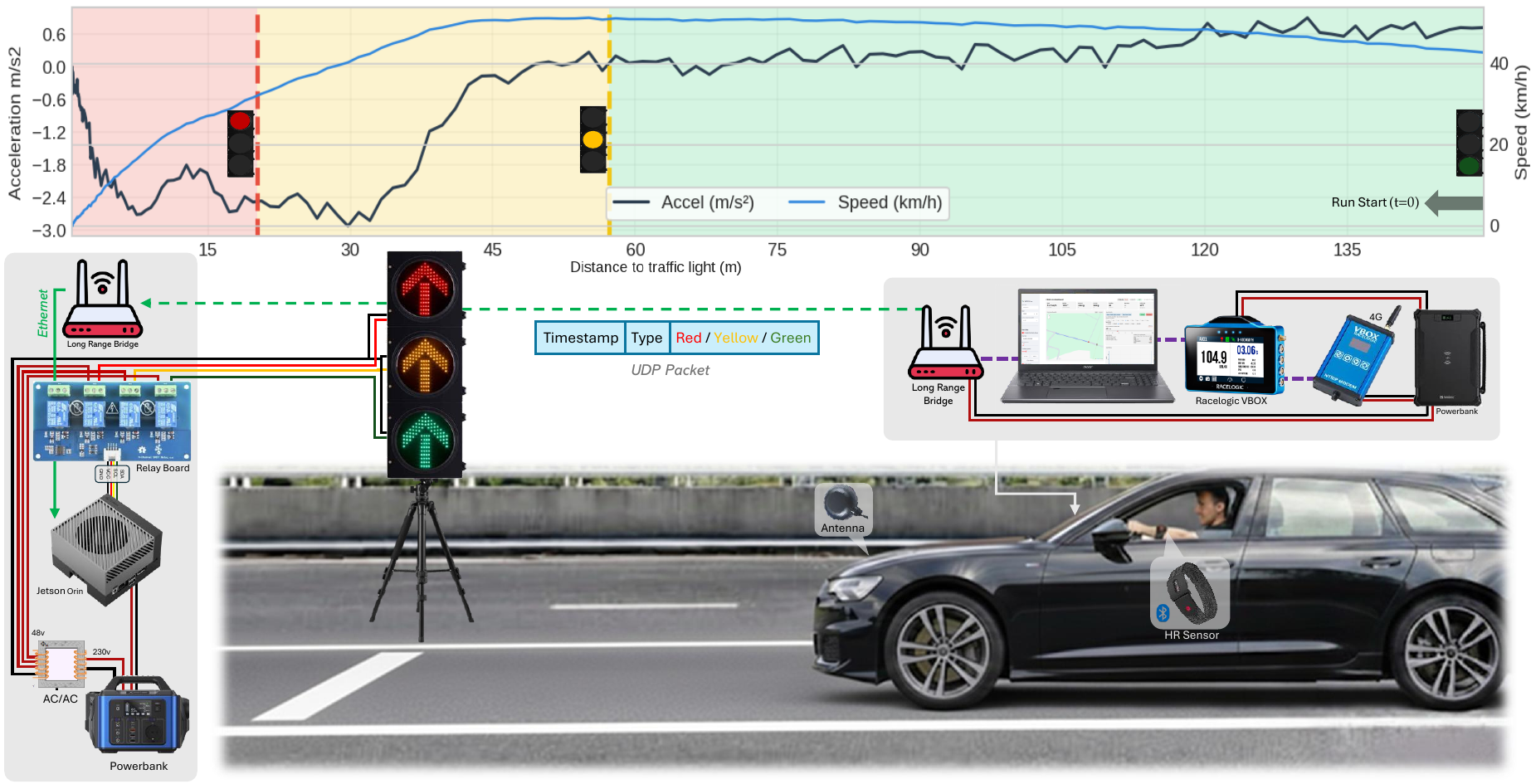}
    \caption{Overview of the experimental configuration. The figure shows the roadside traffic light unit, vehicle-side sensing and communication system, and the wireless data exchange architecture. The top panel illustrates a representative stopping run with longitudinal acceleration and speed profiles versus distance to the traffic light.}
    \label{fig:setup_overview}
\end{figure}

Our contributions address these challenges in three directions. To the best of our knowledge, this is the first real-world dataset for driver trajectories at signalized intersections that combines centimeter-level vehicle motion with synchronized driver physiology and stopping-comfort feedback (Fig.~\ref{fig:setup_overview}). The data were collected using a production vehicle with RTK-corrected GNSS and a dedicated software system that controls traffic light timing from vehicle distance and dynamics while synchronously logging all signals. We statistically analyze driver behavior to identify the main factors governing decision and braking dynamics and use these findings to design an efficient generative model. We propose a two-stage decision-conditioned autoregressive (DCAR) Transformer \cite{vaswani2017attention} that predicts the maneuver decision and then generates the full deceleration trajectory until stopping or crossing. Physics-based constraints enforce realistic acceleration profiles. Experiments across multiple configurations show that a single yellow-onset snapshot is sufficient to generate the complete stopping trajectory and estimate driver comfort level.

\section{Related Work}
\label{sec:related_work}

Comfort-oriented deceleration has been moderately explored in current driving literature. Studies suggest passengers tolerate longitudinal decelerations around $-3.5$ m/s$^2$ \cite{carlowitz2026balancing}, with $-4$ m/s$^2$ acting as an acceptable comfort limit during automated collision avoidance \cite{wang2015improved}. Despite these findings, establishing quantified, multi-level comfort profiles explicitly for red-light decelerations remains an unaddressed gap.

Driver behavior at signalized intersections has been studied from both decision and trajectory perspectives. Early field studies statistically described deceleration rates and stopping probabilities at yellow onset but did not perform predictive modeling \cite{el2007evaluation, el2011impact}. Stop/go behavior has been studied with binary choice and probabilistic regression models relating approach speed, distance to the stop line, and arrival time to stopping probability and violation risk \cite{caird2007effect, papaioannou2007driver, li2016predicting, hurwitz2012fuzzy, masoud2024machine}. Early regression frameworks provide interpretable decision boundaries but often rely on camera-based or roadside sensing, which lacks the spatial and temporal precision of RTK-GNSS and typically omits detailed driver demographics. Machine learning models using support vector machines, random forests, adaptive boosting, and classification trees improve predictive accuracy and capture nonlinear kinematic interactions \cite{jahangiri2015adopting, elhenawy2015modeling, elmitiny2010classification}, but remain limited to discrete decision prediction and do not generate continuous deceleration trajectories. Stochastic and state-transition approaches add temporal evolution through Monte Carlo simulations of required acceleration in dilemma zones \cite{amer2011agent}, hazard-based conflict probability during yellow onset \cite{sharma2011estimating}, and hidden Markov inference of future driver states from sequential speed and signal data \cite{li2016predicting}. While these models incorporate time structure, most do not explicitly represent the cognitive dynamics of decision formation. Recent work emphasizes the continuous nature of yellow-phase decisions \cite{biswas2020reliability} and broader behavior characterization using interpretable machine learning for driving-style classification \cite{ma2023modeling} or structural equation methods for scrambling behavior \cite{qi2021modeling}, although combining decision logic with physically consistent deceleration-trajectory generation remains underexplored.

While datasets like SIND \cite{xu2022drone} use drones to capture valuable vehicle dynamics, they lack extended tracking distances, driver-specific features, and the centimeter-level accuracy of RTK-GNSS. This lack of precision is a critical limitation, as the GNSS-only accuracy that current V2I systems rely upon is insufficient for high-fidelity tracking, particularly in urban environments where buildings further degrade signal quality \cite{campolo20245gnss}. Furthermore, although simulation-based trajectory estimation \cite{calvi2022stop} can generate diverse scenarios with rich participant data, human subjects within these virtual environments cannot physically experience the actual forces of deceleration, nor can they feel the true sensation of speed found in a real-world vehicle.

Therefore, a critical need remains for a dedicated, high-precision dataset at signalized intersections that incorporates comprehensive participant profiles, physiological responses, and multi-level comfort ratings during stopping to address this gap. Furthermore, applying autoregressive transformer architectures \cite{vaswani2017attention} to this precise and comprehensive data leads to a more realistic estimation of vehicle trajectories and human driver decisions.

\section{Experimental Setup}
\label{sec:experiment}

\subsection{Data Collection}

The study captures three coupled aspects of yellow onset behavior under realistic driving conditions. These are the stop go decision at yellow onset, the resulting vehicle trajectory, and the driver reported acceleration induced stress together with physiological response.

Thirty licensed drivers of different ages and both genders participated. Each completed the full protocol individually under identical conditions. Data collection lasted eight days. In each run participants drove along a straight traffic-free road longer than 300~m and approached a single instrumented traffic light. More than 80~m of unobstructed road was available after the traffic light so that surrounding traffic did not influence or distract the driver during the stop/go decision. Before the main trials, participants completed a familiarization phase of about 5--10 minutes to adapt to brake sensitivity and vehicle control. One or two practice runs followed to confirm task understanding. All received identical written instructions and were asked to follow normal traffic rules. A green-only run was randomly inserted to reduce anticipation bias and was excluded from dataset. All experiments were conducted during daytime at the same location using the same vehicle. Each participant performed twelve main runs at instructed speeds of 40, 50, or 60~km/h. Selected participants additionally completed 70~km/h runs as well. Drivers were told that deviations of approximately $\pm 5$~km/h were acceptable to eliminate the distraction effect. This produced a continuous distribution of actual yellow onset speeds ($v_{\mathrm{onset}}$), which is more suitable for behavior modeling than discrete nominal speeds. 
In addition to voluntary decision runs, two forced-stop runs were performed at the end of each session. Participants were instructed to stop regardless of their intention. These runs captured higher deceleration magnitudes that occur less frequently in natural decisions. After each stopping maneuver, drivers rated the stress induced by the applied acceleration on a five point scale where 1 indicates very high stress and 5 indicates very comfortable and smooth deceleration.

The traffic signal automatically switched from green to yellow at predefined upstream distance thresholds ($D_{\mathrm{thr}}$) between 12~m and 79~m. Before the main study two authors conducted repeated pilot drives to identify speed distance combinations near the hesitation region between stop and go. The final design increased sampling density around this region while still including shorter and longer distances that typically lead to clear decisions.

Vehicle position, speed, longitudinal acceleration, and traffic signal state were recorded during each run. Heart rate was continuously measured with a wearable sensor. Recording continued for about 12~s after standstill to capture delayed cardiovascular response.

\subsection{Vehicle, Sensor, and Traffic Light Setup}

Experiments were conducted using a standard passenger vehicle (Audi A6, automatic, diesel) to ensure consistent braking characteristics and avoid drivetrain-dependent effects such as regenerative braking. A portable three-aspect traffic light was installed roadside and controlled via a relay interface connected to an NVIDIA Jetson Orin. The Jetson received UDP commands from the vehicle-side system and executed signal transitions. Vehicle–infrastructure communication used outdoor wireless access points (TP-Link EAP215) to provide stable long-range, low-latency connectivity. Vehicle kinematics (position, speed, longitudinal acceleration) were recorded using a Racelogic VBOX Touch data logger \cite{racelogic_vbox_touch} with RTK correction via NTRIP modem. Runs were initiated only after RTK fix. RTK provided centimeter-level accuracy compared to meter-scale standalone GNSS errors. The GNSS antenna was hood-mounted with a metallic ground plane. Data were logged at 10~Hz and streamed to a laptop. Heart rate was measured using a Polar Verity Sense sensor (transferred via BLE to the laptop). A custom software interface developed by the authors managed real-time data acquisition, traffic light control, calibration, and experiment configuration. Green was activated at run start. Yellow was triggered automatically when the vehicle reached predefined approach distances via UDP commands to the roadside controller. To replicate typical European speed-dependent signal timing, red followed after $3\,\mathrm{s}$ ($\leq 50\,\mathrm{km/h}$), $4\,\mathrm{s}$ ($60\,\mathrm{km/h}$), and $5\,\mathrm{s}$ ($70\,\mathrm{km/h}$) of yellow.

\subsection{Distance Calibration and Longitudinal Correction}
\label{sec:distance_calibration}

Let the GNSS measurement at time \(t\) be \(\mathbf{p}(t)=[\varphi(t),\lambda(t),h(t)]^\top\) in geodetic coordinates. Let \(\mathbf{p}_{\mathrm{TL}}\) denote the traffic light reference location and \(\mathbf{p}_0\) a reference point placed 200~m upstream on the road centerline. Because vehicles may not travel exactly along the centerline, direct geodetic distance introduces lateral bias. The longitudinal distance is therefore obtained by projection onto the road axis defined by the segment connecting \(\mathbf{p}_0\) and \(\mathbf{p}_{\mathrm{TL}}\).

Geodetic coordinates are mapped to a local East–North frame centered at \(\mathbf{p}_{\mathrm{TL}}\). With mean Earth radius \(R\), first-order linearization yields
\begin{equation}
\Pi(\mathbf{p}(t)) =
\begin{bmatrix}
R \cos(\varphi_{\mathrm{TL}})\big(\lambda(t)-\lambda_{\mathrm{TL}}\big) \\
R \big(\varphi(t)-\varphi_{\mathrm{TL}}\big)
\end{bmatrix},
\label{eq:enu_mapping}
\end{equation}
where angular differences are expressed in radians.

Let the road direction be given by \(\Pi(\mathbf{p}_0)\). The signed longitudinal distance to the traffic light is
\begin{equation}
d(t)=
\frac{\Pi(\mathbf{p}_0)^\top \Pi(\mathbf{p}(t))}{\|\Pi(\mathbf{p}_0)\|_2},
\label{eq:distance_projection}
\end{equation}
which is positive upstream, zero at the reference, and negative after passing the traffic light. The projection removes lateral-offset bias in direct geodetic distance.

In addition to spatial calibration, temporal latency in the sensing and actuation chain shifts the effective yellow onset position. If this delay is ignored, the visible yellow transition occurs at a shorter physical distance than intended. Measurements with high-speed video and visual road markers show that the resulting spatial offset can reach up to 4.5~m depending on vehicle speed. The total latency includes relay switching, I2C and Jetson execution, optical rise time of the traffic light, and internal buffering delay of the GNSS logger. The combined fixed latency of these components was measured on average as \(\tau_{\mathrm{lat}} = 142~\mathrm{ms}\). Additionally, a communication latency between the vehicle and the Jetson unit is considered and measured at each timestamp via a ping-based function \(L_{\mathrm{ping}}(t_k)\).

At each discrete update instant \(t_k\), the software predicts the remaining time to the desired yellow onset distance \(D_{\mathrm{thr}}\) from the current corrected longitudinal distance \(d(t_k)\) and speed \(v(t_k)\). The yellow command send time is computed as
\begin{equation}
t_{\mathrm{send}}(t_k)
=
t_k
+
\max\!\left(
0,\,
\frac{d(t_k)-D_{\mathrm{thr}}}{\max\!\big(v(t_k),\epsilon_v\big)}
-
\big(\tau_{\mathrm{lat}}+L_{\mathrm{ping}}(t_k)\big)
\right),
\label{eq:send_time}
\end{equation}
where \(\epsilon_v>0\) prevents numerical instability at very low speed. If the predicted time-to-threshold is smaller than the latency budget, the command is sent immediately. Otherwise, the transmission is delayed accordingly. This computation is repeated at each update, and the command is issued after the last feasible instant to reduce sensitivity to speed fluctuations. After applying this calibration procedure, the spatial onset error was reduced to below 30~cm. Further reduction was limited by the unknown and time-varying internal latency of the GNSS logger in the current setup.

\begin{figure}[t]
\vspace{0.15 cm}
  \centering
  \includegraphics[width=\columnwidth]{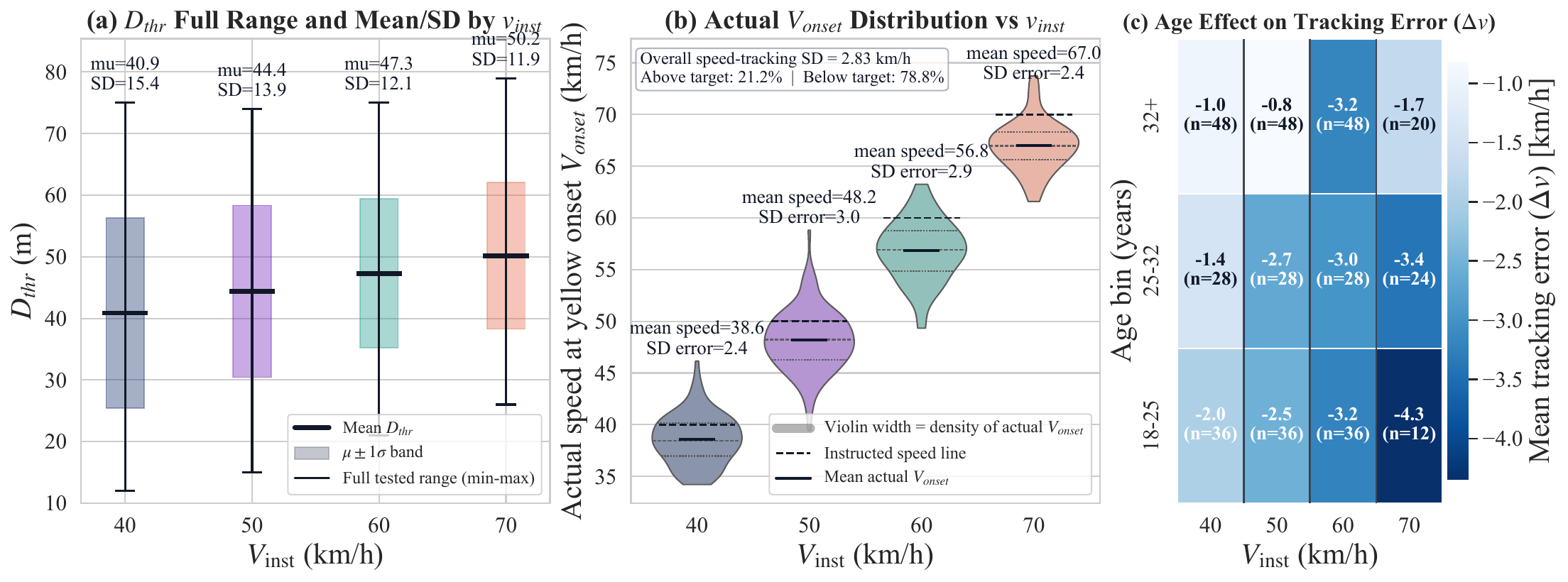}
  \caption{Summary of experiment-setting and speed-tracking characteristics across instructed-speed conditions. (a) Yellow-onset distance-threshold selection \(D_{\mathrm{thr}}\) versus instructed speed \(v_{\mathrm{inst}}\), shown with mean\(\pm\)SD. (b) Actual speed at yellow onset \(v_{onset}\) for each \(v_{\mathrm{inst}}\), with instructed-speed references and compact tracking statistics (mean speed and speed-error SD). (c) Age-binned mean underspeed magnitude \((\Delta v = v_{\mathrm{inst}}-v_{onset})\) versus \(v_{\mathrm{inst}}\), shown in km/h (raw values, darker blue indicates larger underspeed).}
  \label{fig:dt_speed_tracking}
\end{figure}

\section{Dataset Characteristics and Statistical Structure}
\subsection{Run Distribution and Speed Tracking}

A total of 449 runs were recorded (185{,}109 GNSS samples), comprising 392 normal runs for stop–go analysis and the remainder dedicated to high-deceleration forced-stop evaluation.

Figure~\ref{fig:dt_speed_tracking}(a) summarizes yellow-onset thresholds $D_{\mathrm{thr}}$ by instructed speed $v_{\mathrm{inst}}$. Thresholds span 12–79~m with broad coverage across all speed groups. The mean $D_{\mathrm{thr}}$ increases with $v_{\mathrm{inst}}$, reflecting the experimental design to preserve decision variability at higher speeds. Run allocation is balanced at 40, 50, and 60~km/h ($n=112$ each) and lower at 70~km/h ($n=56$) due to the limited number of realistic high-speed urban scenarios. Figure~\ref{fig:dt_speed_tracking}(b) shows actual yellow-onset speed $v_{\mathrm{onset}}$ measured by RTK-corrected GNSS. In 78.8\% of runs, drivers traveled below the instructed speed. The tracking error $\Delta v = v_{\mathrm{onset}} - v_{\mathrm{inst}}$ equals $-2.26 \pm 2.83$~km/h (mean$\pm$SD). Drivers generally adhered to the instructed speed but maintained a systematic underspeed. This negative bias is consistent with the intentional positive offset in production vehicle speed displays implemented by manufacturers. Figure~\ref{fig:dt_speed_tracking}(c) presents age-binned mean underspeed magnitude $(-\Delta v)$. Underspeed mostly increases with instructed speed, consistent with proportional display bias and greater tracking difficulty at higher velocities. Older participants show descriptively smaller tracking error. However, this age effect is not statistically significant (Kruskal--Wallis $H=3.79$, $p=0.150$).

\subsection{Kinematic Variables}

For each decision run, the longitudinal distance and actual speed at yellow onset are denoted by \(D_{\mathrm{thr}}\) and \(v_{\mathrm{onset}}\). Two derived quantities are defined as

\begin{equation}
\mathrm{TTI} = \frac{D_{\mathrm{thr}}}{v_{\mathrm{onset}}},
\qquad
a_{\mathrm{req}} = \frac{v_{\mathrm{onset}}^2}{2 D_{\mathrm{thr}}},
\label{eq:tti_areq}
\end{equation}

where TTI is the time-to-intersection and \(a_{\mathrm{req}}\) is the constant deceleration required to stop at the traffic light.

\begin{figure}[t]
\vspace{0.15 cm}
  \centering
  \includegraphics[width=\columnwidth]{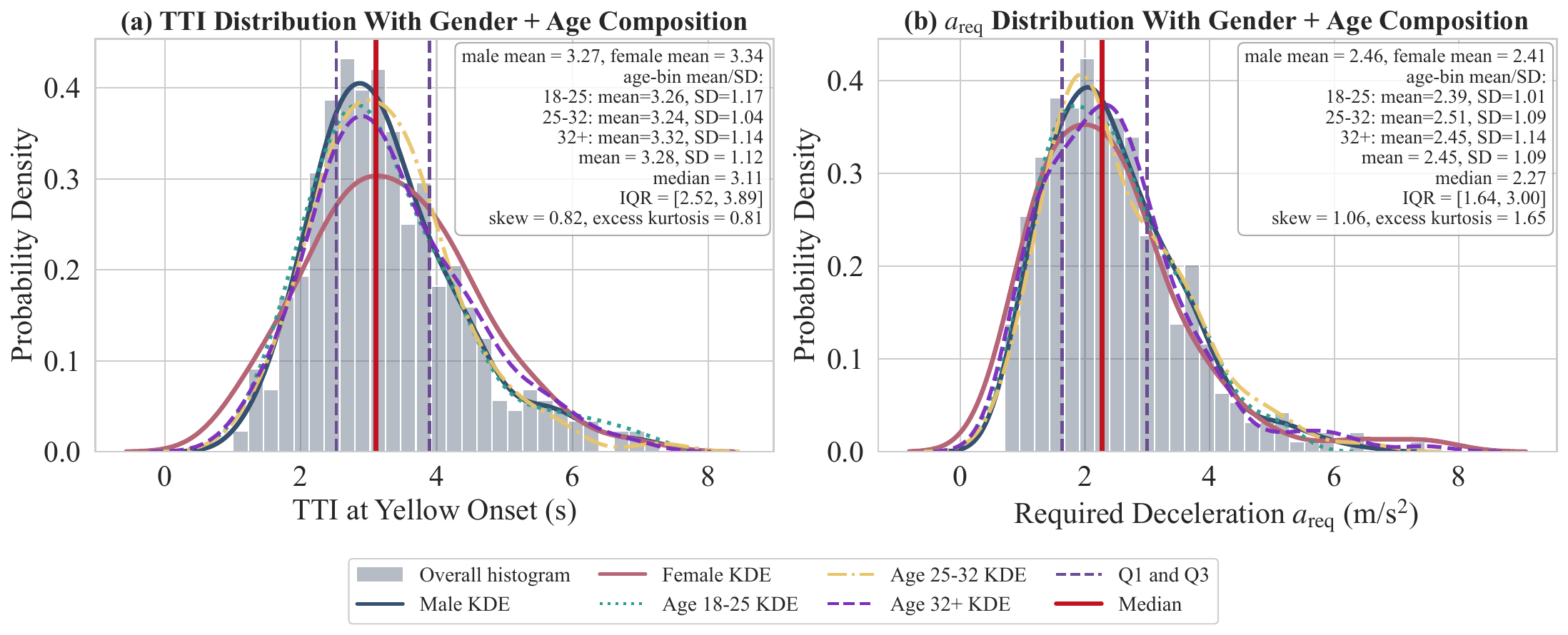}
  \caption{Dataset statistics summary for TTI and required deceleration \(a_{\mathrm{req}}\), including demographic coverage across the observed range.}
  \label{fig:data_collection_stats}
\end{figure}

Figure~\ref{fig:data_collection_stats} presents the empirical distributions of TTI and \(a_{\mathrm{req}}\). The experiment design covers a broad range of both variables, including low-TTI high-deceleration cases and high-TTI moderate-deceleration cases, with sufficient sampling around the transition region. Both genders and all three age groups are represented across the observed span of TTI and \(a_{\mathrm{req}}\). This broad kinematic coverage reduces bias from unequal exposure when comparing demographic groups.

\subsection{Determinants of the Stop--Go Decision}
\label{sec:det_of_stop}

To quantify the influence of kinematic and demographic factors on the stop--go outcome, we employ a Bayesian hierarchical logistic model, whose coefficients and participant-level random intercepts are inferred from data. Let \(y_i \in \{0,1\}\) denote the decision of run \(i\) (1 = go). The model is

\begin{equation}
\left\{
\begin{aligned}
y_i &\sim \mathrm{Bernoulli}(\pi_i), \\
\mathrm{logit}(\pi_i) &= \beta_0 + \sum_{k=1}^{K} \beta_k x_{ik} + u_{p[i]}, \\
u_j &\sim \mathcal{N}(0,\sigma_u^2).
\end{aligned}
\right.
\label{eq:hier_logit}
\end{equation}


As summarized in Table~\ref{tab:ame_compact}, posterior average marginal effects (AMEs) identify \(a_{\mathrm{req}}\) as the dominant stop--go predictor. This indicates that drivers may unconsciously estimate the braking effort required after yellow onset and use this perceived future demand as the main decision cue, beyond observing distance or speed separately.



\begin{figure}[t]
\vspace{0.15 cm}
  \centering
  \includegraphics[width=\columnwidth]{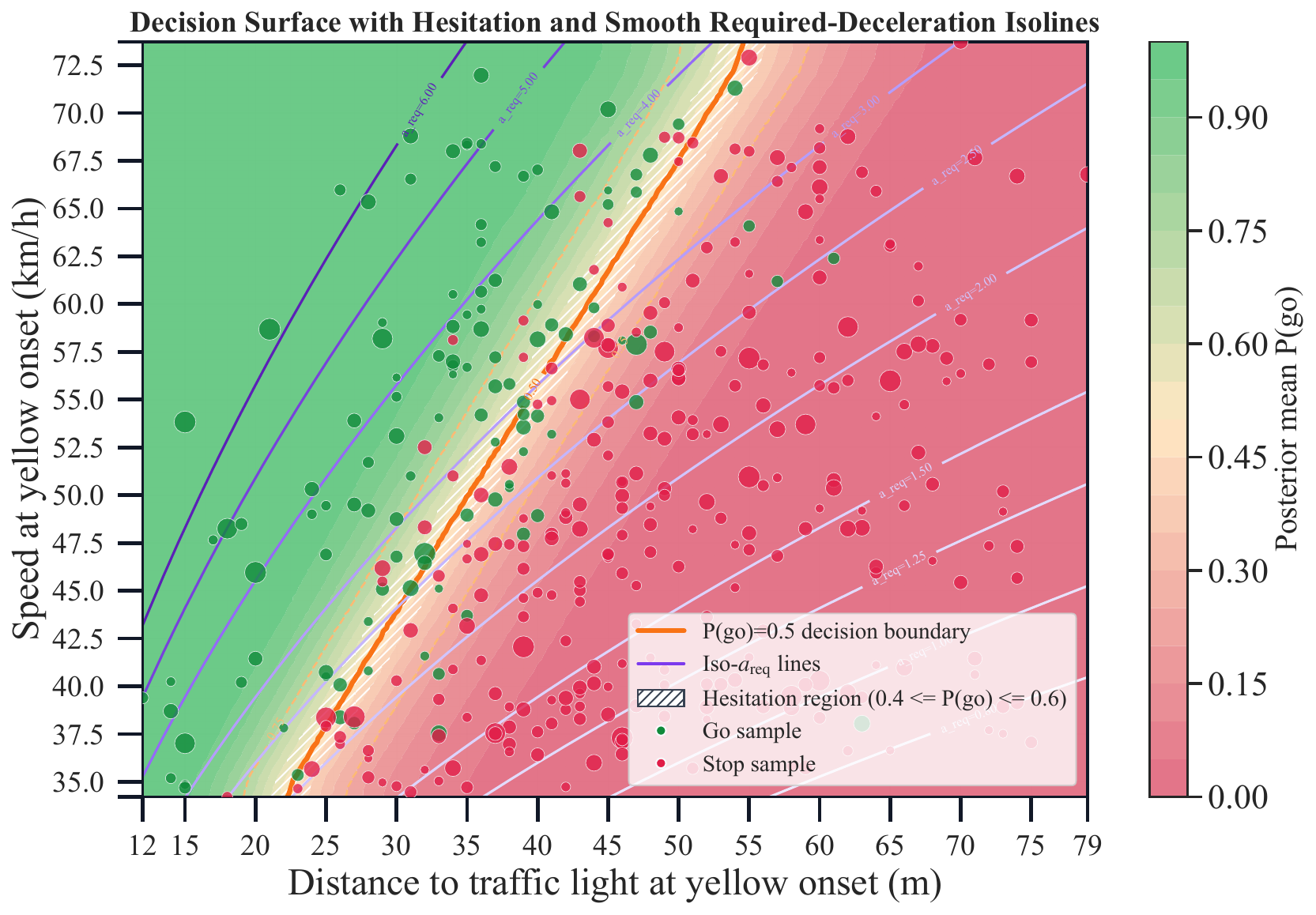}
  \caption{Distance--speed stop/go decision surface at yellow onset. The background colormap shows the posterior mean $P(\mathrm{go})$ from the Bayesian hierarchical logistic model; the thick contour denotes the decision boundary $P(\mathrm{go})=0.5$, and the hatched band marks the hesitation region $0.4 \le P(\mathrm{go}) \le 0.6$. Overlaid sample points are observed runs (green: go, red: stop), with marker size proportional to normalized driver age. Smooth iso-$a_{\mathrm{req}}$ lines are superimposed (light-to-dark purple for low-to-high required deceleration).}

  \label{fig:distance_speed_decision}
\end{figure}

The same hierarchical logistic model is used to evaluate the posterior mean decision surface $\overline{P}_{\mathrm{go}}(D_{\mathrm{thr}}, v_{\mathrm{onset}})$ over the distance--speed plane (Fig.~\ref{fig:distance_speed_decision}). For stop trials ($n=274$), the mean required deceleration (mean \(a_{\mathrm{req}}\)) at yellow onset is $1.99$~m/s$^2$ (median $1.93$, SD $0.70$); the upper 20\% exhibit $a_{\mathrm{req}}$ between $2.58$ and $4.15$~m/s$^2$ (mean $3.05$). This range coincides with the hesitation band, concentrated around $2$--$4$~m/s$^2$.

The decision boundary approximately follows iso-$a_{\mathrm{req}}$ contours but shifts with distance and speed. For larger $D_{\mathrm{thr}}$, the stop–go transition occurs at higher deceleration levels (approximately $3$--$4$~m/s$^2$), whereas at shorter distances it shifts toward lower magnitudes (approximately $2$--$3$~m/s$^2$). This indicates that drivers do not apply a fixed deceleration threshold; instead, braking demand is evaluated jointly with geometric margin. As $a_{\mathrm{req}}$ increases, $\overline{P}_{\mathrm{go}}$ rises sharply, consistent with the AME result. The hesitation band aligns with intermediate-to-high deceleration demand, which suggests that drivers avoid aggressive braking once the maneuver approaches uncomfortable or safety-critical levels.

\begin{table}[t]
\centering
\caption{AME summary for stop--go predictors; AME is in percentage points and \(\mathcal{C}_k\) is relative contribution.}
\label{tab:ame_compact}
\renewcommand{\arraystretch}{1.1}
\setlength{\tabcolsep}{5.6pt}
\small
\begin{tabular}{lcccccc}
\toprule
 & \(D_{\rm thr}\) & \(v_{\rm onset}\) & \(a_{\rm req}\) & TTI & Gender & Age \\
\midrule
AME [pp] & \(-7.5\) & \(-4.3\) & \(\mathbf{+33.7}\) & \(+0.9\) & \(+7.7\) & \(0.0\) \\
\(\mathcal{C}_k\) [\%] & \(14.1\) & \(9.4\) & \(\mathbf{50.8}\) & \(10.2\) & \(12.5\) & \(3.0\) \\
\bottomrule
\end{tabular}
\vspace{-0.2cm}
\end{table}

\subsection{Braking Magnitude, Comfort, and Physiological Response}
\label{sec:comfort}

Peak braking magnitude is defined per run as
\begin{equation}
|a_{\min}| = \max_t \big\{ \max(-a_x(t), 0) \big\},
\end{equation}
using the minimum longitudinal acceleration during braking, expressed as positive deceleration.

\begin{figure}
\vspace{0.15 cm}
    \centering
    \includegraphics[width=1\linewidth]{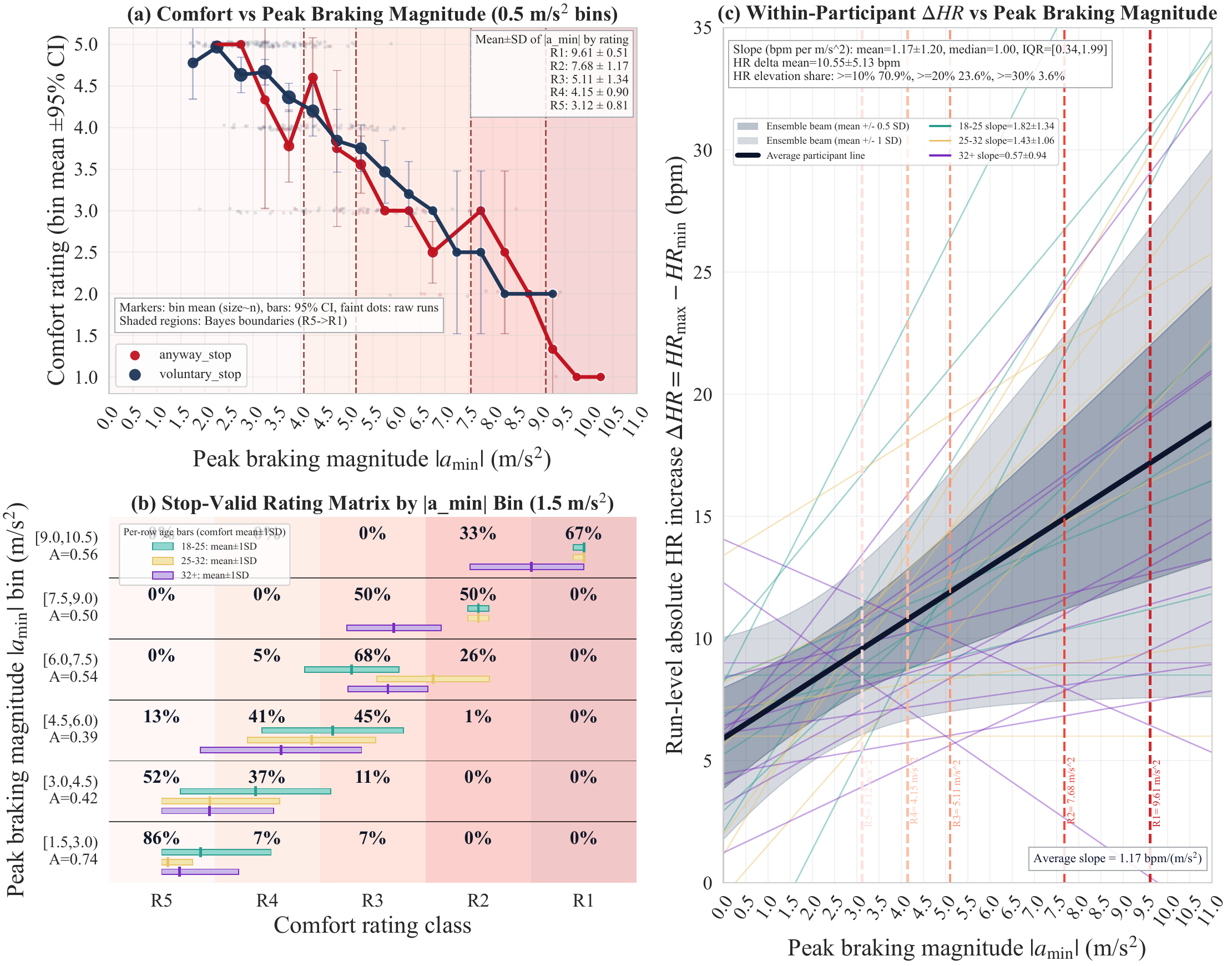}
    \caption{Physiological and comfort responses in stop-valid runs. (a) Comfort rating versus peak braking magnitude \(|a_{\min}|\), with bin-wise means and raw-run overlays. (b) Rating-composition matrix over \(|a_{\min}|\) bins (cell text: within-bin percentage), with age-group bars showing mean comfort \(\pm 1\) SD; row labels report disagreement index \(A=\sum_{k=1}^{5}p_k^2\), where \(p_k\) is the proportion in class \(R_k\) (lower \(A\) indicates higher disagreement). (c) Within-participant relation between \(\Delta HR=HR_{\max}-HR_{\min}\) and \(|a_{\min}|\): thin lines are participant-wise robust trends, shaded bands show ensemble spread (\(\pm0.5\) SD and \(\pm1\) SD), the thick line is the ensemble mean trend, and dashed vertical lines mark rating-specific mean \(|a_{\min}|\) (R5 to R1).}

    \label{fig:HR-comfort}
\end{figure}

Figure~\ref{fig:HR-comfort}(a) shows comfort rating as a function of $|a_{\min}|$. A strong monotonic inverse relationship is observed. Across all stop-valid runs, Spearman correlation yields $\rho=-0.675$ ($p=2.76\times10^{-45}$). In forced-stop runs, the association increases to $\rho=-0.806$ ($p=4.00\times10^{-14}$). The linear trend in forced-stop trials indicates a decrease of approximately $0.483$ rating units per $1$~m/s$^2$, so comfort decreases systematically as braking demand increases. 

To determine rating-transition boundaries on the $|a_{\min}|$ axis, adjacent comfort classes are modeled as heteroscedastic Gaussian densities with empirical priors $\pi_k=n_k/\sum_j n_j$. Because classes exhibit unequal variances and imbalanced sample sizes, simple midpoint thresholds or equal-variance linear discriminant rules would be statistically inconsistent; therefore, boundaries are derived using a prior-weighted heteroscedastic Gaussian Bayes criterion.

The Bayes-optimal boundary $x_{ij}^\star$ between classes $R_i$ and $R_j$ satisfies equality of log-posteriors,
\begin{equation}
-\frac{(x_{ij}^\star-\mu_i)^2}{2\sigma_i^2}
-\ln\sigma_i
+\ln\pi_i
=
-\frac{(x_{ij}^\star-\mu_j)^2}{2\sigma_j^2}
-\ln\sigma_j
+\ln\pi_j,
\label{eq:comfort_boundary}
\end{equation}
which yields a quadratic equation in $x_{ij}^\star$. When two real solutions exist, the threshold consistent with the ordered rating scale is selected; if no crossing occurs strictly between adjacent means under empirical priors, the nearest real crossing preserving monotone class ordering is used. Applying this rule results in transition boundaries of $4.07$, $5.15$, $7.54$, and $9.11$~m/s$^2$ for $(R5|R4)$, $(R4|R3)$, $(R3|R2)$, and $(R2|R1)$, respectively.

Figure~\ref{fig:HR-comfort}(b) presents rating composition across deceleration bins and quantifies agreement via
\begin{equation}
A \;=\;
\sum_{k=1}^{5}\!\left(\frac{n_k}{\sum_{j=1}^{5} n_j}\right)^{\!2},
\end{equation}
where $n_k$ denotes the count of ratings in class $R_k$ within a deceleration bin. Higher $A$ indicates stronger concentration and lower disagreement. The highest agreement occurs in the low-deceleration interval $[1.5,3.0)$~m/s$^2$ ($A=0.745$), whereas the lowest agreement appears in $[4.5,6.0)$~m/s$^2$ ($A=0.386$). The intermediate braking regime therefore exhibits the largest perceptual dispersion, consistent with the absence of a well-separated posterior crossing between adjacent rating densities in this range and the resulting outward-shifted Bayes boundary.

Age-group overlays show that younger drivers generally assign lower comfort ratings for the same braking magnitude. The only exception appears in the $[6.0,7.5)$~m/s$^2$ interval, where the ordering reverses, likely due to differing internal reference frames rather than sampling effects.

Figure~\ref{fig:HR-comfort}(c) shows the within-participant relation between heart rate elevation, defined as the difference between maximum and minimum heart rate during a stopping run, and $|a_{\min}|$. Participant-wise robust slopes yield an overall mean of 1.173 bpm per $1$~m/s$^2$. Age-stratified slopes decrease systematically with age, indicating reduced physiological sensitivity to braking in older drivers. This aligns with the comfort ratings, where older participants consistently report higher comfort at comparable deceleration levels. Greater driving experience and age therefore appear to be associated with higher tolerance to braking-induced stress.

\section{Modeling Human Deceleration Behavior}
\label{sec:modeling}

\subsection{Training Setup}

All experiments primarily use a random stratified split (60\% train, 40\% validation), with additional evaluation under a participant-disjoint split. Models are implemented in Python~3.8.20 with PyTorch~1.10.1 (CUDA~11.1) and trained on a single NVIDIA GeForce RTX~4090 Laptop GPU.

Stage~1 (decision MLP) is trained for 200 epochs using Adam (LR $10^{-3}$, batch size 64, hidden size 64, weight decay $10^{-5}$). Stage~2 (DCAR Transformer) uses cascade-only conditioning from Stage~1 with true-masking enabled and is trained for 200 epochs using AdamW (LR $7\times10^{-4}$, batch size 24, weight decay $10^{-5}$, gradient clipping 4.0).

\begin{figure*}
\vspace{0.15 cm}
    \centering
    \includegraphics[width=1\linewidth]{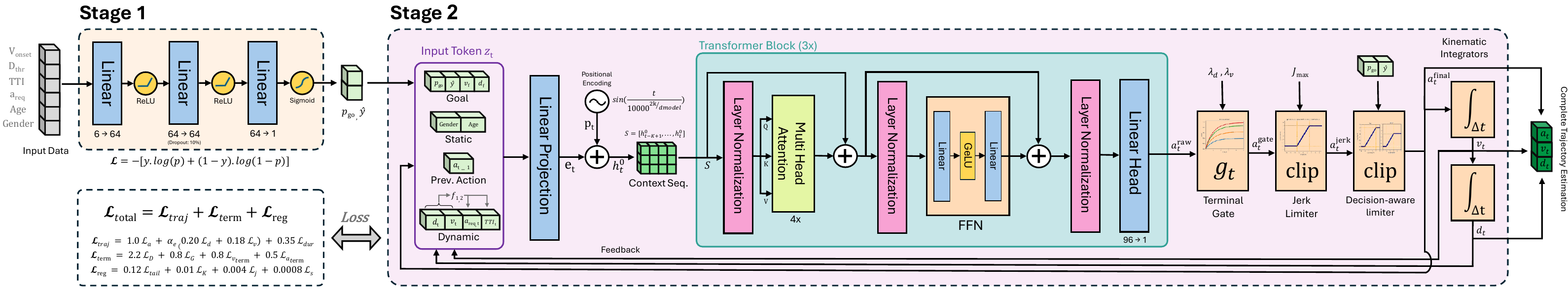}
    \caption{Two-stage decision-conditioned autoregressive trajectory generator with deterministic constraint pipeline and discrete vehicle dynamics feedback.}
    \label{fig:model_overview}
\end{figure*}

\subsection{Two-Stage Trajectory Generator}
\label{sec:two_stage_ar_transformer}

We model yellow-onset behavior using a two-stage decision-conditioned autoregressive architecture. A discrete maneuver decision is predicted once at onset and then held fixed while a causal Transformer generates a physically constrained longitudinal acceleration trajectory in closed loop (Fig.~\ref{fig:model_overview}).

\subsubsection{Stage 1: Maneuver Decision Modeling}

An onset-level MLP maps $[V_{\mathrm{onset}}, D_{\mathrm{thr}}, \mathrm{TTI}, a_{\mathrm{req}}, \mathrm{age}, g]$ to $p_{\mathrm{go}}\in(0,1)$ and $\hat y=\mathbb{I}[p_{\mathrm{go}}\ge0.5]$, where $g$ denotes gender. The network consists of three fully connected layers with ReLU activations and 10\% dropout, followed by a Sigmoid output layer. It is trained using class-weighted binary cross entropy. The outputs $(\hat y,p_{\mathrm{go}})$ condition the full trajectory rollout in Stage~2.

\subsubsection{Stage 2: Autoregressive Trajectory Modeling}
The Transformer \cite{vaswani2017attention} predicts signed longitudinal
acceleration stepwise as $a_t^{\mathrm{raw}}=\mathcal{T}_\theta(\mathbf{z}_{\le t})$,
where the causal token at step $t$ is
\[
\mathbf{z}_t =
\big[
v_t,\ d_t,\ a_{\mathrm{req},t},\ \mathrm{TTI}_t,\ a_{t-1},\ \mathrm{age},\
\hat y,\ p_{\mathrm{go}},\ g,\ v_{\text{target}},\ d_{\text{target}}
\big].
\]
Here $(v_{\text{target}}=0, d_{\text{target}}=0.5)$ encodes the goal state for stopping cases, and $a_{t-1}$ denotes the previously applied (post-constraint) acceleration.
Tokens are embedded as $\mathbf{e}_t=\mathbf{W}_e\mathbf{z}_t+\mathbf{p}_t$ with
$\mathbf{z}_t\in\mathbb{R}^{11}$, \(d_{\mathrm{model}}=96\), four heads, and \(d_k=24\).

Causal multi-head self-attention is
\begin{equation}
\operatorname{Attn}(\mathbf{Q},\mathbf{K},\mathbf{V})
=
\operatorname{softmax}
\!\left(
\frac{\mathbf{Q}\mathbf{K}^{\!\top}+\mathbf{M}}{\sqrt{d_k}}
\right)\mathbf{V},\;\;
M_{ij}=\left\{\begin{array}{@{}l@{\,}l@{}}
0 & j\leq i\\
-\infty & j>i
\end{array}\right.
\end{equation}
ensuring strictly autoregressive rollout.
The output projection head produces $a_t^{\mathrm{raw}}$, which is processed by a deterministic constraint pipeline before integration.
A decision-dependent terminal gate attenuates acceleration near terminal
conditions,
\begin{equation}
a_t^{\mathrm{gate}}=g_t\,a_t^{\mathrm{raw}},
\qquad
g_t=
\left(1-e^{-d_t^{\mathrm{safe}}/\lambda_d}\right)
\left(1-e^{-v_t^{\mathrm{safe}}/\lambda_v}\right),
\end{equation}
where $v_t^{\mathrm{safe}}=\max(v_t,0)$ and
\[
d_t^{\mathrm{safe}}=
\begin{cases}
\max(d_t+0.5,0), & \hat y=0,\\
\max(d_t+0.25,0), & \hat y=1,
\end{cases}
\]
with $(\lambda_d,\lambda_v)=(2.2~\mathrm{m},1.2~\mathrm{m/s})$ for stop trajectories and $(0.8~\mathrm{m},1.5~\mathrm{m/s})$ for go trajectories.

A hard jerk constraint is then applied using the discrete time step $\Delta t$ and $J_{\max}=15~\mathrm{m/s^3}$,
\begin{equation}
a_t^{\mathrm{jerk}}
=
\operatorname{clip}
\!\big(
a_t^{\mathrm{gate}},
\ a_{t-1}-J_{\max}\Delta t,\ 
a_{t-1}+J_{\max}\Delta t
\big),
\end{equation}
enforcing bounded acceleration rate changes for rollout stability. The jerk cap is selected based on empirical driver data, where the maximum observed absolute jerk was $|j|_{\max}=12.69~\mathrm{m/s^3}$; choosing $J_{\max}=15~\mathrm{m/s^3}$ provides a small margin above the observed range while remaining consistent with realistic vehicle dynamics.

A decision-aware actuator clamp enforces feasible control ranges
($a_{\min}^{\mathrm{brake}}=-10~\mathrm{m/s^2}$,
$a_{\max}^{\mathrm{go}}=3~\mathrm{m/s^2}$),
\begin{equation}
a_t=
\begin{cases}
\operatorname{clip}(a_t^{\mathrm{jerk}},a_{\min}^{\mathrm{brake}},0),
& \hat y=0,\\[4pt]
\operatorname{clip}(a_t^{\mathrm{jerk}},a_{\min}^{\mathrm{brake}},a_{\max}^{\mathrm{go}}),
& \hat y=1,
\end{cases}
\end{equation}
preventing positive acceleration in stop trajectories while allowing bounded acceleration in go trajectories. The constrained acceleration is integrated with discrete kinematics,
\begin{equation}
\begin{aligned}
d_{t+1}&=d_t-\left(v_t\Delta t+\tfrac{1}{2}a_t\Delta t^2\right),\\
v_{t+1}&=\max(v_t+a_t\Delta t,0),
\end{aligned}
\end{equation}
and a terminal freeze is enforced once the trajectory enters a small stopping deadband (where stop-band success is defined as a final position within $[-0.5,\,1.5]$~m relative to the stop line) or reaches near-zero velocity, after which $a_t=0$ and $v_t=0$ are maintained. The model is trained on the post-constraint trajectory so that gating and clipping behavior are internalized during learning.

The training objective combines trajectory fidelity, terminal feasibility, and smoothness regularization,
\[
\mathcal{L}
=
\mathcal{L}_{\mathrm{traj}}
+
\mathcal{L}_{\mathrm{term}}
+
\mathcal{L}_{\mathrm{reg}},
\]
ensuring accurate trajectory tracking while enforcing physically consistent terminal behavior.


\section{Results}
\label{sec:results}

Table~\ref{tab:traj_model_compare} compares four trajectory-generation approaches on the stratified split. The DCAR Transformer achieves the lowest acceleration and trajectory-tracking errors among the learned models across all $\varepsilon$-metrics. The constant deceleration baseline enforces a fixed braking profile $a(t)=-a_{\mathrm{req}}$ for stop cases and therefore attains 100\% stop-band success deterministically. However, this guarantees terminal feasibility only and results in substantially larger trajectory-tracking errors. The mixture-of-primitives reduces error relative to the constant model but remains clearly inferior to the DCAR Transformer. Moreover, based on the comfort boundaries derived in Section~\ref{sec:comfort}, the DCAR Transformer generates braking profiles that match the ground-truth driver comfort level in 84.5\% of stop cases. This shows that future driver stopping comfort can be anticipated from a single observation at yellow onset.

\begin{table}[t]
\vspace{0.15 cm}
\centering
\caption{Trajectory-model comparison on the stratified split. Lower is better for \(\varepsilon\)-metrics; higher is better for \(\rho_{\mathrm{stop}}^{\mathrm{band}}\)(\%).}
\label{tab:traj_model_compare}
\setlength{\tabcolsep}{1.7pt}
\tiny
\resizebox{0.91\columnwidth}{!}{%
\begin{tabular}{l | c c c c | c}
\toprule
Method &
\(\varepsilon_{a(\cdot)}\) & \(\varepsilon_{a(\cdot)}\) & \(\varepsilon_{d(\cdot)}\) & \(\varepsilon_{v(\cdot)}\) &
\(\rho_{\mathrm{stop}}^{\mathrm{band}}\) \\
& MAE & RMSE & MAE & MAE & success \\
\midrule
\(a(t)=-a_{\mathrm{req}}\,[1-y]\)
& 0.865 & 1.170 & 1.404 & 0.854 & \textbf{100.0} \\
Mixture of primitives
& 0.856 & 1.331 & 1.470 & 1.781 & 90.0 \\
Neural ODE
& 1.058 & 1.686 & 1.447 & 1.715 & 75.5 \\
DCAR Transformer w/o physics
& 0.559 & 1.032 & 0.666 & 0.426 & 98.2 \\
DCAR Transformer (\textbf{ours})
& \textbf{0.499} & \textbf{0.850} & \textbf{0.628} & \textbf{0.373} & 95.5 \\
\bottomrule
\end{tabular}%
}
\end{table}

Table~\ref{tab:stopgo_ablation_expanded} reports Stage~1 stop/go ablation under random and participant-disjoint splits. The strongest compact configuration is $(V_{\mathrm{onset}}, a_{\mathrm{req}})$, which achieves the highest accuracy (92.4\%) and the strongest F1 and precision in both splits among all tested combinations. This indicates that onset speed and required deceleration already capture most of the discriminative structure required for stop/go classification. Among single-cue models, \(a_{\mathrm{req}}\) and TTI are strongest, and \(D_{\mathrm{thr}}\) exceeds \(v_{\mathrm{onset}}\), mostly consistent with Section~\ref{sec:det_of_stop}. The full-input model gives the highest random-split AUC and remains competitive under participant-disjoint evaluation.

Under the participant-disjoint split, overall performance decreases relative to the random split. Since participant ID is never used as input in either setting, this gap indicates that the model can no longer benefit from implicit memorization of driver-specific decision boundaries.

\begin{table}[t]
\centering
\caption{Stage~1 stop/go ablation (by onset features) under random (stratified run-level) and participant-disjoint splits. All metric entries are percentages (\%), where $\eta_{\mathrm{acc}}$ and $\eta_{\mathrm{BA}}$ denote accuracy and balanced accuracy, respectively. Bold values denote the highest values in each metric column.}
\label{tab:stopgo_ablation_expanded}
\setlength{\tabcolsep}{1.6pt}
\tiny
\resizebox{\columnwidth}{!}{%
\begin{tabular}{c c c c c c | c c c c c | c c c c c}
\toprule
\multirow{3}{*}{\rotatebox{90}{\(V_{\mathrm{onset}}\)}} &
\multirow{3}{*}{\rotatebox{90}{\(D_{\mathrm{thr}}\)}} &
\multirow{3}{*}{\rotatebox{90}{TTI}} &
\multirow{3}{*}{\rotatebox{90}{\(a_{\mathrm{req}}\)}} &
\multirow{3}{*}{\rotatebox{90}{Age}} &
\multirow{3}{*}{\rotatebox{90}{Gender}} &
\multicolumn{5}{c}{Random (stratified) split} &
\multicolumn{5}{c}{Participant-disjoint split} \\
\cmidrule(lr){7-11}\cmidrule(lr){12-16}
 &  &  &  &  &
 & \(\eta_{\mathrm{acc}}\) & AUC & F1 & Prec. & \(\eta_{\mathrm{BA}}\)
 & \(\eta_{\mathrm{acc}}\) & AUC & F1 & Prec. & \(\eta_{\mathrm{BA}}\) \\
\midrule
\checkmark & $\times$     & $\times$     & $\times$     & $\times$     & $\times$     & 59.9 & 69.4 & 54.0 & 41.1 & 65.3 & 60.4 & 64.7 & 44.9 & 34.8 & 61.3 \\
$\times$   & \checkmark   & $\times$     & $\times$     & $\times$     & $\times$     & 79.0 & 84.8 & 69.7 & 61.3 & 79.5 & 79.2 & 89.2 & 69.3 & 55.6 & 83.4 \\
$\times$   & $\times$     & \checkmark   & $\times$     & $\times$     & $\times$     & 91.1 & 95.8 & 86.5 & 78.9 & 92.4 & 83.2 & 97.8 & 75.2 & 60.3 & 88.7 \\
$\times$   & $\times$     & $\times$     & \checkmark   & $\times$     & $\times$     & 84.1 & 93.0 & 77.5 & 67.2 & 86.2 & 83.9 & 95.2 & 75.0 & 62.1 & 87.5 \\
\checkmark & \checkmark   & $\times$     & $\times$     & $\times$     & $\times$     & 89.8 & 96.2 & 84.3 & 78.2 & 90.3 & 84.6 & 98.2 & 76.8 & 62.3 & 89.6 \\
\checkmark & $\times$     & \checkmark   & $\times$     & $\times$     & $\times$     & 90.4 & 96.2 & 85.4 & 78.6 & 91.4 & 84.6 & 98.0 & 76.8 & 62.3 & 89.6 \\
\checkmark & $\times$     & $\times$     & \checkmark   & $\times$     & $\times$     & \textbf{92.4} & 96.4 & \textbf{88.0} & \textbf{83.0} & \textbf{92.7} & \textbf{87.9} & 97.9 & \textbf{80.4} & \textbf{68.5} & 91.0 \\
$\times$   & \checkmark   & \checkmark   & $\times$     & $\times$     & $\times$     & 90.4 & 96.0 & 85.4 & 78.6 & 91.4 & 84.6 & 98.1 & 76.8 & 62.3 & 89.6 \\
$\times$   & \checkmark   & $\times$     & \checkmark   & $\times$     & $\times$     & 91.1 & 96.1 & 86.5 & 78.9 & 92.4 & 84.6 & 98.1 & 76.8 & 62.3 & 89.6 \\
$\times$   & $\times$     & \checkmark   & \checkmark   & $\times$     & $\times$     & 90.4 & 96.2 & 85.4 & 78.6 & 91.4 & 84.6 & 98.1 & 76.8 & 62.3 & 89.6 \\
\checkmark & $\times$     & $\times$     & \checkmark   & \checkmark   & $\times$     & 90.4 & 95.7 & 84.8 & 80.8 & 90.1 & 85.9 & 97.2 & 77.9 & 64.9 & 89.7 \\
\checkmark & $\times$     & $\times$     & \checkmark   & $\times$     & \checkmark   & 91.1 & 96.4 & 85.7 & 82.4 & 90.6 & 85.9 & 96.7 & 77.9 & 64.9 & 89.7 \\
\checkmark & \checkmark   & \checkmark   & $\times$     & $\times$     & $\times$     & 90.4 & 96.3 & 85.4 & 78.6 & 91.4 & 83.9 & \textbf{98.3} & 76.0 & 61.3 & 89.2 \\
\checkmark & \checkmark   & $\times$     & \checkmark   & $\times$     & $\times$     & 89.8 & 95.8 & 84.3 & 78.2 & 90.3 & 83.9 & 98.1 & 76.0 & 61.3 & 89.2 \\
\checkmark & $\times$     & $\times$     & \checkmark   & \checkmark   & \checkmark   & 89.2 & 96.5 & 83.2 & 77.8 & 89.2 & 85.2 & 97.5 & 77.6 & 63.3 & 90.1 \\
\checkmark & \checkmark   & \checkmark   & \checkmark   & $\times$     & $\times$     & 90.4 & 96.6 & 85.4 & 78.6 & 91.4 & 84.6 & 98.2 & 76.8 & 62.3 & 89.6 \\
\checkmark & \checkmark   & \checkmark   & \checkmark   & $\times$     & \checkmark   & 89.8 & 96.6 & 84.0 & 79.2 & 89.7 & 87.2 & 97.3 & 80.0 & 66.7 & \textbf{91.4} \\
\checkmark & \checkmark   & \checkmark   & \checkmark   & \checkmark   & $\times$     & 89.2 & 95.2 & 83.2 & 77.8 & 89.2 & 83.9 & 98.0 & 76.0 & 61.3 & 89.2 \\
\checkmark & \checkmark   & \checkmark   & \checkmark   & \checkmark   & \checkmark   & 90.4 & \textbf{96.7} & 85.4 & 78.6 & 91.4 & 83.9 & 98.1 & 76.0 & 61.3 & 89.2 \\
\bottomrule
\end{tabular}%
}
\end{table}

Table~\ref{tab:ar_ablation_splitA} details Stage~2 input ablations using the optimal Stage~1 setup. The best performance occurs when using all inputs except age, which minimizes all primary tracking and terminal errors ($\varepsilon_{d_f}$, $\varepsilon_{v_f}$). Notably, ablating the Stage~1 decision increases acceleration error by up to 2.6$\times$, demonstrating the transformer's inability to isolate driver intention independently and validating our two-stage approach. Stop-band success peaks under two configurations, including the full model.

\begin{table}[t]
\vspace{0.2 cm}
\centering
\caption{Stage~2 input-modality ablation on the stratified split. All \(\varepsilon\)-metrics are MAE values in corresponding physical units. Lower is better for \(\varepsilon\)-metrics and higher is better for \(\rho_{\mathrm{stop-band}}\). Bold values denote the best values columns.}
\label{tab:ar_ablation_splitA}
\setlength{\tabcolsep}{1.5pt}
\tiny
\resizebox{\columnwidth}{!}{%
\begin{tabular}{c c c c c c c | c c c c c c}
\toprule
\multicolumn{7}{c|}{Input features} & \multicolumn{6}{c}{Validation metrics} \\
\cmidrule(lr){1-7}\cmidrule(lr){8-13}
\rotatebox{0}{\(V_t\)} &
\rotatebox{0}{\(D_t\)} &
\rotatebox{0}{TTI} &
\rotatebox{0}{\(a_{req}\)} &
\rotatebox{0}{Age} &
\rotatebox{0}{\(\hat y\)} &
\rotatebox{0}{\(p_{go}\)}
& \(\varepsilon_{a(\cdot)}\) & \(\varepsilon_{v(\cdot)}\) & \(\varepsilon_{d(\cdot)}\) & \(\varepsilon_{d_f}\) & \(\varepsilon_{v_f}\) & \(\rho_{\mathrm{stop-band}}\) \\
\midrule
$\checkmark$ & $\checkmark$ & $\times$     & $\times$     & $\times$     & $\times$     & $\times$     & 1.15 & 0.80 & 1.15 & 1.54 & 1.53 & 94.55 \\
$\checkmark$ & $\checkmark$ & $\times$     & $\checkmark$ & $\times$     & $\times$     & $\times$     & 1.33 & 0.89 & 1.23 & 1.44 & 1.65 & 96.36 \\
$\times$     & $\times$     & $\checkmark$ & $\checkmark$ & $\checkmark$ & $\times$     & $\times$     & 1.34 & 1.01 & 1.33 & 1.70 & 1.63 & \textbf{100.00} \\
$\checkmark$ & $\checkmark$ & $\checkmark$ & $\checkmark$ & $\times$     & $\times$     & $\times$     & 0.96 & 0.91 & 2.42 & 3.13 & 1.70 & 69.09 \\
$\times$     & $\times$     & $\checkmark$ & $\checkmark$ & $\times$     & $\checkmark$ & $\times$     & 0.51 & 0.41 & 0.76 & 0.98 & 0.45 & 99.09 \\
$\checkmark$ & $\checkmark$ & $\checkmark$ & $\checkmark$ & $\checkmark$ & $\times$     & $\times$     & 1.31 & 0.84 & 0.84 & 1.41 & 1.67 & 99.09 \\
$\checkmark$ & $\checkmark$ & $\checkmark$ & $\times$     & $\checkmark$ & $\checkmark$ & $\times$     & 0.51 & 0.55 & 1.27 & 1.41 & 0.51 & 98.18 \\
$\checkmark$ & $\checkmark$ & $\times$     & $\times$     & $\checkmark$ & $\checkmark$ & $\checkmark$ & 0.51 & 0.55 & 1.27 & 1.35 & 0.51 & 99.09 \\
$\checkmark$ & $\checkmark$ & $\checkmark$ & $\checkmark$ & $\checkmark$ & $\checkmark$ & $\times$     & 0.52 & 0.41 & 0.67 & \textbf{0.89} & 0.45 & 99.09 \\
$\checkmark$ & $\checkmark$ & $\checkmark$ & $\checkmark$ & $\times$     & $\checkmark$ & $\checkmark$ & \textbf{0.49} & \textbf{0.37} & \textbf{0.62} & 0.92 & \textbf{0.39} & 95.45 \\
$\checkmark$ & $\checkmark$ & $\checkmark$ & $\checkmark$ & $\checkmark$ & $\checkmark$ & $\checkmark$ & 0.51 & 0.43 & 0.75 & 0.91 & 0.45 & \textbf{100.00} \\
\bottomrule
\end{tabular}%
}
\end{table}

\begin{figure}[t]
    \centering
    \includegraphics[width=\columnwidth]{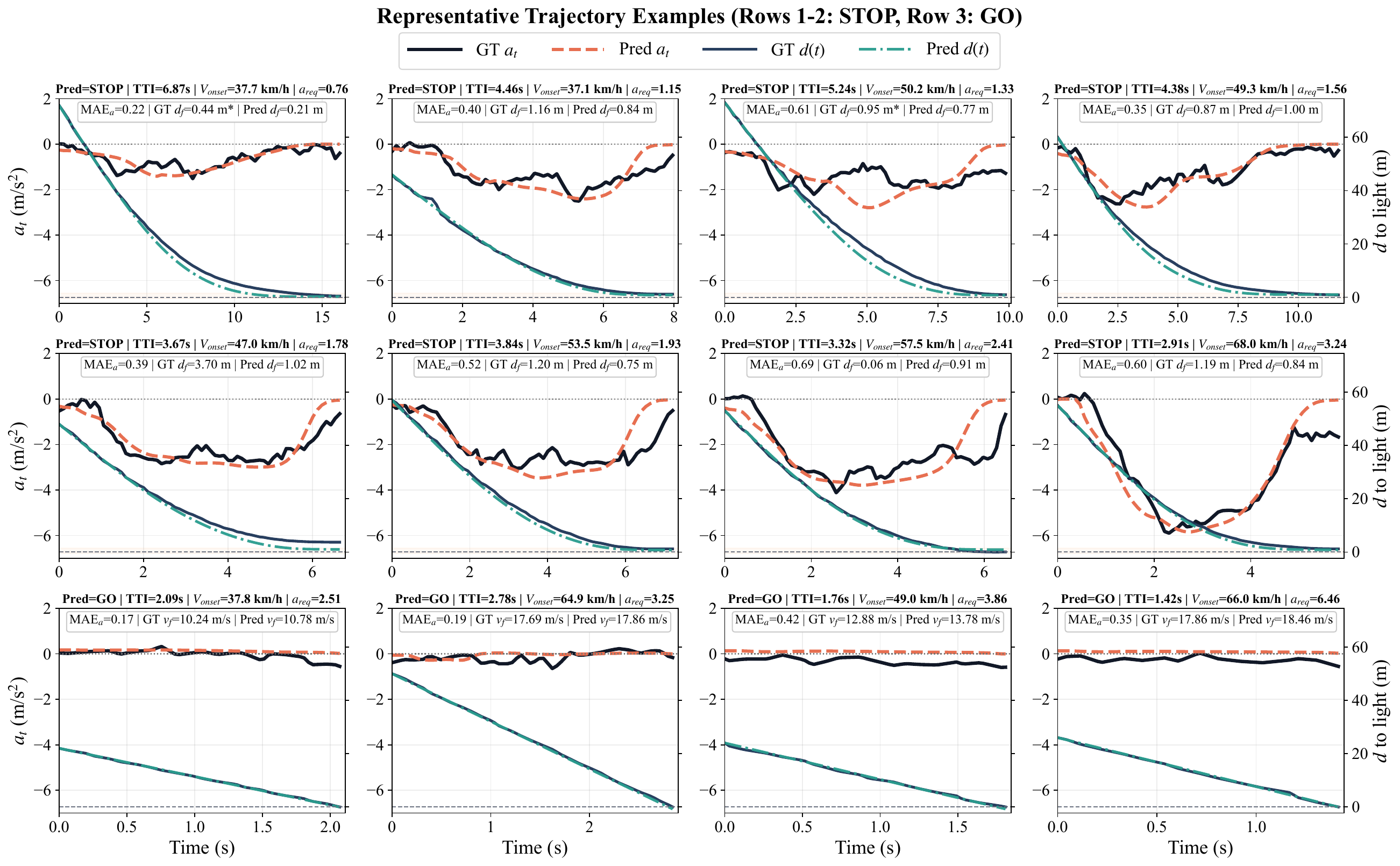}
    \caption{Twelve trajectory estimation examples (rows 1--2 stop and row 3 go). Each panel shows predicted and measured acceleration and distance trajectories in physical time with panel-level TTI, \(v_{onset}\), \(a_{\mathrm{req}}\), acceleration MAE, and outcome annotations \(d_f\) for stop and \(v_f\) for go.}
    \label{fig:ar_traj_examples}
\end{figure}

Figure~\ref{fig:ar_traj_examples} supports these findings qualitatively. Predicted accelerations are smoother because of jerk regularization while preserving the dominant two-stage human-like stop pattern. Stop examples keep low positive \(d_{f}\) values, which means final stopping positions remain before and close to the stop line in most cases.

\section{Discussion and Outlook}
\label{sec:discussion}

Compared with camera-based data collection approaches, our study uses an RTK-GNSS-based measurement setup that provides substantially higher positioning accuracy and enables precise correlation of multi-level acceleration-induced comfort with each stopping trajectory. To further increase scenario diversity, the dataset can be extended with additional contextual factors such as the presence and distance of a following vehicle, pedestrians at the crosswalk, and variations in lighting and weather conditions.
The current dataset is collected in Germany and primarily includes drivers holding German driving licenses. Driving habits, stopping strategies, and comfort perception may differ across countries and traffic cultures, particularly outside Europe. Cross-region data collection is therefore important for evaluating model generalization and behavioral variability.
Another relevant direction is the study of passenger comfort during stopping maneuvers. Since passengers do not control the vehicle, perceived braking discomfort and physiological response may differ substantially from the driver's perspective. Incorporating passenger-side measurements could improve the design of human-centered automated braking systems.

\section{Conclusion}
\label{sec:conclusion}

We developed a high-precision experimental setup in which traffic light timing is triggered from RTK-corrected vehicle position. This allows accurate and repeatable yellow-onset generation. Using this setup, we collected a real-world dataset containing vehicle kinematics, braking-comfort ratings, and physiological responses from human drivers. Based on this dataset, we proposed a two-stage decision-conditioned autoregressive Transformer that predicts the stop/go decision and generates the full longitudinal trajectory from a single yellow-onset observation. The model achieves an acceleration MAE of $0.49~\mathrm{m/s^2}$ and a distance-trajectory MAE of $0.62~\mathrm{m}$ while maintaining high stop-band success with accurate final stopping positions.

The learned trajectories additionally allow estimation of future braking stress from onset-level kinematics alone. The results indicate that both maneuver intent and comfort-related braking behavior can be predicted early and support predictive intersection safety systems and infrastructure-assisted driver support.

\section*{Acknowledgment}

This work was supported by the KIT Future Fields Wild Ideas project ``WildRobot'', the Helmholtz program Engineering Digital Futures (EDF), and the ELLIIT Excellence Center. The authors gratefully acknowledge the support of Stadtmobil CarSharing GmbH, Karlsruhe.


\bibliographystyle{IEEEtran} 
\bibliography{ref}

\end{document}